\documentclass[conference]{IEEEtran}

\usepackage[dvips]{graphicx}
\usepackage{amsmath,amsfonts,amssymb}
\usepackage{algorithm}
\usepackage{algorithmic}
\usepackage{url}        
\usepackage{amsmath}    
\usepackage{hyperref}
\usepackage{tikz}
\usepackage{caption}
\usepackage{subcaption}

\begin{document}
\title{Towards Blockchain-based Multi-Agent Robotic Systems: Analysis, Classification and Applications 
}
\date{}

\author{
\IEEEauthorblockN{Ilya Afanasyev\textsuperscript{1}, Alexander Kolotov\textsuperscript{1}, Ruslan Rezin\textsuperscript{1}, Konstantin Danilov\textsuperscript{1}, Manuel Mazzara\textsuperscript{1} \\ Subham Chakraborty\textsuperscript{1}, Alexey Kashevnik\textsuperscript{2}, Andrey Chechulin\textsuperscript{2}, Aleksandr  
Kapitonov\textsuperscript{2} \\ Vladimir Jotsov\textsuperscript{3}, Andon  Topalov\textsuperscript{4}, Nikola Shakev\textsuperscript{4}, Sevil Ahmed\textsuperscript{4}
}
\IEEEauthorblockA{\textsuperscript{1}Innopolis University, Innopolis, Russia \\
\textsuperscript{2}ITMO University, St.Petersburg, Russia \\
\textsuperscript{3}University of Library Studies and Information, Sofia, Bulgaria \\
\textsuperscript{4}Technical University of Sofia, Branch Plovdiv, Plovdiv, Bulgaria \\
\{i.afanasyev, a.kolotov, r.rezin, k.danilov, m.mazzara, s.chakraborty\}@innopolis.ru \\ alexey@iias.spb.su, chechulin@comsec.spb.ru, kapitonov.aleksandr@itmo.ru \\ v.jotsov@unibit.bg,  \{topalov, shakev, sevil.ahmed\}@tu-plovdiv.bg\\
}
}

\maketitle

\begin{abstract}
Decentralization, immutability and transparency make of Blockchain one of the most innovative technology of recent years. This paper presents an overview of solutions based on Blockchain technology for multi-agent robotic systems, and provide an analysis and classification of this emerging field. The reasons for implementing Blockchain in a multi-robot network may be to increase the interaction efficiency between agents by providing more trusted information exchange, reaching a consensus in trustless conditions, assessing robot productivity or detecting performance problems, identifying intruders, allocating plans and tasks, deploying distributed solutions and joint missions. Blockchain-based applications are discussed to demonstrate how distributed ledger can be used to extend the number of research platforms and libraries for multi-agent robotic systems.
\end{abstract}
\section{Introduction}
\label{introduction}

The successes demonstrated in recent years in the integration of robotic systems, wireless sensor networks (WSN), cloud computing, distributed planning and management, and distributed ledgers provides and optimistic outlook towards increasingly popular technological solutions such as the Internet of Robotic Things (IoRT) \cite{ray2016internet, simoens2018internet, batth2018internet, mahieu2019semantics, afanasyev2019towards} and the Blockchain-based Multi-Agent Robotic Systems (MARS) \cite{kapitonov2017, ferrer2018blockchain, strobel2018managing, afanasyev2019blockchain}.
~
It is known that one of the important problems in developing multi-robot systems is the design of strategies for their coordination in such a way that the robots could effectively perform their operations and reasonably coordinate the task allocation among themselves \cite{basegio2017decentralised}. Real-world scenarios usually require the use of heterogeneous robots and the performance of tasks with various structures, constraints and complexity. The task distribution for decentralized solutions is appropriate, since the use of autonomous multi-robot systems in complex scenarios becomes limited and inefficient, and centralized solutions pose a danger of failure for the entire system. Since system agents have to share information, the requirements for the quality of communication in decentralized systems are increasing, including such important functions as maintaining data integrity, resiliency and security in accessing data. Therefore, implementation of blockchain technology for interaction and coordination of multi-agent robotic systems become a reasonable solution for a dynamic and decentralized task distribution.

Studies of distributed ledgers demonstrate that decentralization and immutable record technologies make blockchain one of the most powerful innovations, since in a decentralized network, intercepting most network nodes using cyberattacks looks economically impractical \cite{edChain}. However, many popular blockchain solutions suffer from such issues as scalability, delay and low throughput \cite{edChain}. For example, Bitcoin can process less than 10 TPS (transactions per second), while Ethereum can process up to 40 TPS, which is clearly not enough compared to daily 2000 TPS VISA (which theoretically can increase up to ~50,000 TPS). Since the blockchain network grows with an increase in both the number of users and transactions, a verification of transactions slows down the transaction process and throughput. This is known as the classical Blockchain Trilemma - when it comes to the choice two of the three between decentralization, scalability and security \cite{coinbureau2019}. One of the scaling methods that does not compromise security or decentralization is called “sharding”, which involves fragmentation of the available dataset into smaller datasets called “shards” \cite{edChain, coinbureau2019}.
Although multi-agent robotic systems (MARS) are not so critical to scalability and speed as the financial and big data-based systems, they are nevertheless also very sensitive to delays and throughput of the information channels at data exchange between agents.

The literature describes many types of developed algorithms for distributed consensus, each of which has distinctive features, advantages and disadvantages. The summary of the most important distributed consensus algorithms, including Proof of Work (PoW), Proof of Stake (PoS), Proof of Activity (PoAc), Proof of Burn (PoB), Proof of Capacity (PoC), etc. are presented in the review \cite{andoni2019blockchain}. The methodologies used to achieve consensus in blockchain networks largely determine key performance characteristics, including scalability, transaction speed, transaction completeness, security, and resource consumption. Each method requires a procedure to generate and then adopt a block. The block can be generated or offered by a node in the network, and it encodes a number of transactions (for example, in cryptocurrency, it can be monetary transactions between accounts). Further, a key step is the adoption of the proposed block/related transactions by network participants, a process called consensus building. Once a block is accepted, it becomes part of the block chain, and the newly created blocks are cryptographically linked to it. After some time (depending on the consensus algorithm used), the block becomes a constant part of the blockchain, i.e. reaching a finality. However, the finality does not exclude the existence of a small statistical probability that a block can be changed (intentionally by design or due to an attack), although with each new block added, and for an established blockchain system, it becomes negligible.

Let's summarize the key points of the blockchain technology in relation to the work of multi-agent robotic systems \cite{afanasyev2019blockchain}:
\begin{itemize}
\item The blockchain-based database is a database for adding only. Once the data is included in the database, they cannot be changed. The database forms the blockchain state and is distributed among the nodes.
\item Each node is an agent on the blockchain network and stores a complete copy of the database. The node is responsible for transferring all incoming data received from another node to all its neighboring nodes and could generate records for changing the blockchain state. All nodes are connected through peer-to-peer communication channels. Some nodes should play the role of a validator.
\item Validators check the correctness of records for changing the state of the blockchain and approve them (for example, combining the records into blocks, linking the blocks together and sending new blocks to the neighbors). Only verified entries are applied to all nodes to build the current state of the blockchain.
\end{itemize}


Let's emphasize the advantages of the blockchain technology for multi-agent systems \cite{afanasyev2019blockchain}: 
\begin{itemize}
\item Data availability is achieved through multiply duplication of data and communication.
\item Consistency of data is achieved through data validation and strict rules of changes appliance. 
\item No way to remove or change the data stored in the blockchain.
\item Economic or reputational incentive forces nodes to not violate the validation rules. 
\end{itemize}


The research relevance is based on the importance in the development of distributed multi-agent robotic systems that could effectively perform different operations and independently coordinate the task allocation within the system. Information exchange during the interaction of robots has the particular importance for reaching the goals of a multi-agent system in conditions of uncertainty, external interference, environmental changes or the presence of intruders when data integrity maintenance, resiliency and security have special value. To this end, the blockchain technology for multi-agent robotic systems is designed to solve the problems of information exchange for a group of robots, provide a record of the interaction history and validate the task execution, enhancing the efficiency of the whole system and extending the capabilities of MARS applications.


This paper extends the study presented by the authors in the paper \cite{afanasyev2019blockchain} with new materials related to blockchain-based multi-agent systems, including aspects of implementing MARS via Wireless Sensor Network (WSN), ensuring the integrity and security using distibuted ledgers, and deployment prospects for these systems in Smart Buildings, Smart Cities and Industry 4.0. 

The continuation of this paper is structured as follows. Section \ref{related_papers} introduces the present state of scientific and engineering development in the blockchain-based multi-agent systems. Section \ref{classification} describes and classifies the most typical cases, which we identified for blockchain-based robotics applications. Section \ref{wsn} considers aspects of MARS realization using Wireless Sensor Network. Section \ref{smart_cities} discusses multi-agent robotic systems related to Smart Buildings, Smart Cities and Industry 4.0. Finally, we summarize the strong and weak sides of the blockchain-based MARS and discuss the barriers that technology must overcome in order to prove its viability and become mass in the Section \ref{discussion}. 

\section{Related Papers}
\label{related_papers}

In this section, we analyze the state-of-the-art publications and applications where blockchain technology is used for distributed multi-agent systems with the special focusing on robotics.
~
A recent research series focuses on the use of Blockchain technology for the shared knowledge and reputation management system in studying the collective behavior of robots \cite{kapitonov2017, ferrer2018blockchain, strobel2018managing, zikratov2016dynamic, danilov2018,  teslya2018, Kashevnik2018}.

The study \cite{zikratov2016dynamic} presented a trust management model for decentralized robotic networks that focuses on access control and reputation management for each node. This model provides group access based on a robot-oriented trust that is selected and dynamically updated over time. The analysis of the system was carried out by compromising a robot using attacks.
~
The idea of using blockchain technology to solve security problems in multi-robot systems were discussed in \cite{ferrer2018blockchain, danilov2018}. The author \cite{ferrer2018blockchain} states that combining peer-to-peer networks with cryptographic algorithms allows reaching an agreement by a group of agents (with the following recording this agreement in a verifiable manner) without the need for a controlling authority. He describes some blockchain-based innovations that could provide a breakthrough in MARS applications:
\begin{itemize}
    \item New security models and methods to preserve data confidentiality and robot's entity validation;
    \item Design of distributed decision making and collaborative missions using special transactions in the ledger that allow robotic agents to vote and reach agreements;
    \item Development of blockchain ledgers for using different robot's parameters, corresponding to changing environments without any changes in their control algorithm, allowing to increase the flexibility of robots without increasing the complexity of MARS design;
    \item Creation of infrastructure for MARS to follow certain legal norms and safety rules adopted for human society that could even result in building new business models for MARS operation.
\end{itemize}
~
In the paper \cite{strobel2018managing}, the theoretical concept of managing security problems in multi-robot systems using blockchain technology was reinforced by the implementation and proof-of-concept for controlling Byzantine robots. The authors developed an approach to using decentralized programs based on smart contracts to create secure swarm coordination mechanisms, as well as for identifying and eliminating Byzantine swarm members through collective decision making.


~
The study \cite{kapitonov2017} is based on the organization of the blockchain protocol for multi-agent coordination and control of unmanned aerial vehicles (UAVs).
~
The paper \cite{danilov2018} concerns the consensus protocol of the blockchain, which uses an additional procedure for verifying the liability execution to prevent payment transactions to questionable service providers. For this purpose, the liability execution for agent-based service providers in the decentralized trading market is verified by a formal model checker. As the proof-of-concept, an application was implemented, where a taxi was modeled with the subsequent delivery check at the end of the completed mission.
~
The article \cite{lopes2019} proposes a modular architecture, combining the RobotChain \cite{ferrer2018robochain} framework as a decentralized ledger for registering events with robots, smart contract technology for managing robots and Oracle for processing any data types.
The modular architecture can be used in various contexts (manufacturing, network or robot management, etc.) since it is easy to integrate, adapt, maintain, and expand for new domains. What is more, this architecture allows to refuse from tokens to accelerate the validation process or replace them by a reputation system for managing tasks and reaching consensus, since the monetary value may not make sense for private blockchain-based networks.
The examples of robotic applications can be: 
\begin{itemize}
\item Task allocation between robot network; 
\item Information support of robots in operations (for example, a robot cannot recognize objects, while others can); 
\item Assessing the robot productivity or detecting performance problems; 
\item Voting consensus for swarm robotics. 
\end{itemize}

To ensure the interaction of heterogeneous robots in the cyber-physical space, an ontology can be used that describes the knowledge and competencies of the robots in the system, provides a quick exchange of information between coalition members and smart contracts for the allocation of sensory, computational, control and service tasks between intelligent robots, embedded devices and information resources \cite{teslya2018}. To do this, the study \cite{teslya2018} presents a methodology for creating cyber-physical smart space with instructions how to create and manage coalitions of intelligent robots using knowledge processors and information stored in the blockchain.
~
The similar way, the paper \cite{Kashevnik2018} discusses the cyber-physical-social system, which combines smart space technology and blockchain. The interaction between mobile robots and humans is related on ontology-based publication/subscription mechanism, where all data exchange is controlled and key information is stored in the blockchain network.

Intelligent cyber-physical systems can be implemented as multi-agent systems with the ability to schedule tasks by agents \cite{shukla2018}. In such multi-agent systems, the protocol of plan execution should lead to proper completion and optimization of actions, inspite of their distributed execution. However, in unreliable scenarios there is a probability that agents will not follow the protocol due to failures or malicious reasons that result in the plan failure. To prevent such situations, the plan can be executed by agents through smart contracts, ensuring that the task is performed even in an untrusted environment \cite{shukla2018}. Moreover, smart contracts can be automatically generated from manufacturing plans, resulting in automation of the entire system with seamless integration of agents into one cyber-physical system \cite{shukla2018}.
In the similar way, self-sustaining cyber-physical environments are formed in which all critical aspects at both the cyber and physical levels are effectively stimulated, coordinated and supported using blockchain-based mechanisms and protocols for data storage, communication and coordination \cite{Skowronski2019}.
Thus, the advantages of a properly developed blockchain framework for organizing multi-agent systems (which can be extended to both robotic and cyber-physical systems) should meet the following requirements \cite{Skowronski2019}:
\begin{itemize}
\item Decentralization of data storage, which increases the rigidness of the system;
\item Scalability and ease of joining new agents;
\item Participants reach a consensus in conditions without trust;
\item Ability to maintain trust among initially unknown agents;
\item Transparency and immutability;
\item Agents remain fully autonomous, they fully control their identity and private keys;
\item Blockchain data is complete, consistent and accessible.
\end{itemize}
\section{The Classification of the Blockchain-based Robotics Applications}
\label{classification}


In this section, we classify blockchain-based robotic multi-agent systems, which we revealed during the literature review. Let's consider and discuss our vision of possible robotic multi-agent applications based on blockchain technology shown in the Figure 1 (that is the extended version of the classification presented in \cite{afanasyev2019blockchain}).
~
\begin{figure}[!htbp]
    \centering
    \includegraphics[width=\linewidth]{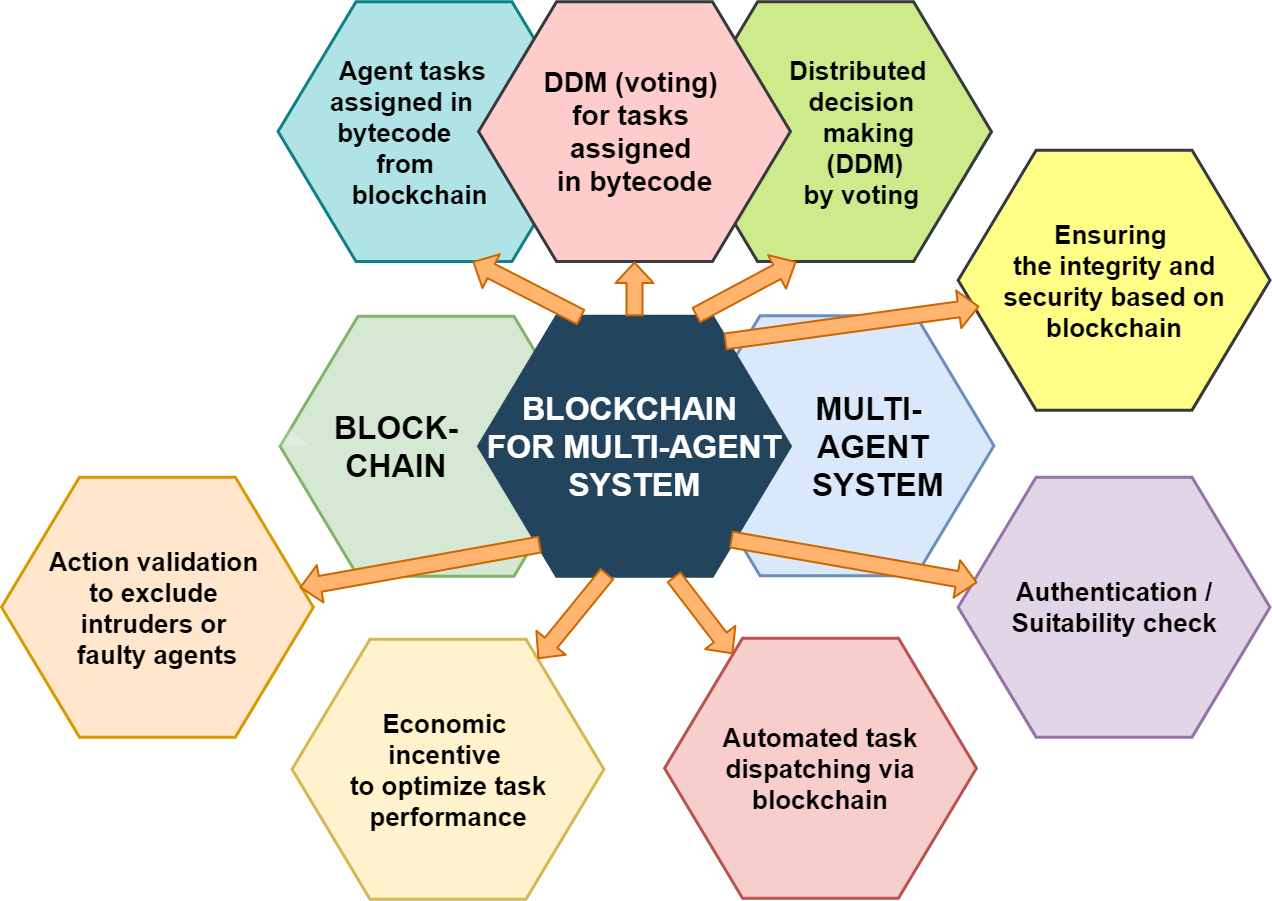}
    \caption{The classification of typical cases for using blockchain technology in multi-agent systems in robotics applications}
    \label{fig01}
\end{figure}

\subsection{Agent Tasks Assigned in Executable Code from Blockchain}

Let's consider a scenario in which there are multiple agents (robots), without taking into account the hardware platform. Any agents have a pre-installed control program, but they are configured to receive external executable code (bytecode), which is a set of commands in the package, to implement operations, achieving the multi-agent system's goal.
~
The word "bytecode" means here the platform-independent executable code that results in the same sequence of command execution for any agents. Thus, at the moment of sending the bytecode, there is no need to know the exact hardware platform of the agent (robot), which must execute the commands specified in the bytecode.

The blockchain could be used here to distribute such bytecodes to the agents \cite{afanasyev2019blockchain}:

- The system which generates the tasks should not be connected to the agent directly. The peer-to-peer network is used to deliver the message.

- A message will be delivered even if an agent is turned off.

- The agent is able to inform constantly about its state changes (e.g. "moved forward for 10 meters", "picked an object", etc.). The state is stored in the blockchain therefore it could be recovered quickly in case of the agent was turned off.

- The delayed bytecode execution could be scheduled.

- Two or more command sequences could be automatically queued for execution by the agent.

\subsection{Distributed Decision Making by a Time-Limited Voting}


Distributed decision making (DDM) is the challenging task in multi-agent robotic systems. There was proposed a solution to use the blockchain in SWARM systems \cite{ferrer2018blockchain}, in which it was used the idea of sending cryptocoins as the prizes to some addresses that must be achieved. However, the development of blockchain technology provides new opportunities that are also effective for solving this problem, for example, using smart contracts in Ethereum. For this purpose, smart contract(s) should be developed to create an infrastructure by conducting polls with complex behaviors, such as time-limited voting or vote delegation.

\subsection{Distributed Decision Making for Tasks Assigned in Bytecode}
By combining the approaches (B) and (A), it is possible to obtain another interesting solution, where the use of smart contracts for agents may contain some tasks formulated in bytecode. Other agents may vote for actions, resulting in defining co-generated script that can be obtained from a smart contract.

\subsection{Action Validation to Exclude Intruders or Faulty Agents}


Agents can be used to check each other’s actions, locations or states. Let's look at the swarm where agents execute some actions to achieve the common goal. Periodically, agents send telemetry information of sensor measurements and their location based on odometry.  Sometimes an agent may start working incorrectly and send wrong data. Information obtained from other agents can be used to reach a consensus that the agent working wrong and the recovery procedure can begin. In this case, co-evolution scenarios can be applied further. A consensus based on information obtained from other agents can also be used to identify a robot that is behaving incorrectly, for example, it had been hacked or infected by an attacker. To solve the performance problem of validators, the Sharding approach can also be used here \cite{edChain, coinbureau2019}. Data from agents are combined depending on their location: thus, the separate shard is formed. Validators are coordinated for the shard, therefore the information volume for processing is significantly reduced.

\subsection{Economic Incentive to Optimize Task Performance}


The financial side of the blockchain can be used as a basis for stimulating a multi-robot system. Thus, researchers from Carnegie Mellon University performed multi-robot mapping with a market approach, coordinating a robot team and maximizing the information acquisition at minimal cost \cite{zlot2002}. This approach demonstrated reliability and adaptability to a dynamic environment, even with the loss of colony members, in addition to its ability to withstand communication losses and disruptions. The researchers found that, the market architecture in the negotiations of robots, improved the robot team efficiency for environmental study in many times \cite{zlot2002}. Although the algorithm was designed to minimize the distance traveled during the mission, minimizing the exploration time also showed encouraging results for quick survey of the terrain. This approach also allows dynamical changing priorities to minimize the resource consumption for the robot team, fulfilling the multiple mission objectives for the robot group \cite{zlot2002}.

\subsection{Automated Task Dispatching via Blockchain}


The blockchain-based distributed consensus can be used to dispatch,  assign and execute tasks between competing agents. In this case, the dispatching code can be written in the form of a smart contract stored in the blockchain for the implementation of the following possible scenario:

\begin{enumerate}
\item The client sends a request for the task execution to the smart contract dispatcher.
\item The dispatcher notifies agents about the new request.
\item The agents agree the task performance in the blockchain via a peer-to-peer network. 
\item Blockchain validators determine the order of agreements in accordance with the commission that a particular agent pays for processing the agreement.
\item The first agreement received by the dispatcher is confirmed by the smart contract code, and the order details are provided to the appropriate agent.
\end{enumerate}



Thus, the market will regulate the choice of agents that will be sent to complete the task. However, certain agents can use strategies that allow them to pay more to be selected validators that may lead to appearing the most efficient and stable service providers. The case study on the creation of such a multi-agent system is presented in the paper \cite{danilov2018}, which implements an automated dispatching taxi script with validation of obligation fulfillment by comparing the route traveled with a map by a validator.

\subsection{Authentication / Suitability Check}

There are critical situations associated with the occurrence of risks / threats that require authentication:
\begin{itemize}
\item When agents do not trust each other, but use a common physical resource;
\item When agents may be attacked by third parties. Thus, hacker attacks can reveal confidential information and/or location, change the task and influence on the agent's activity outcome.
\end{itemize}


In the paper \cite{afanasyev2019blockchain}, the example with a service station for  electric vehicles (EV) battery replacement was considered, where a blockchain-based solution was proposed to provide battery authentication services. Since the smart contract code in the blockchain is not changeable, therefore battery amortization state is available to all participants in the deal. Therefore, the service station and EV can connect to any blockchain nodes to check the battery, excluding data replacement at the man-in-the-middle attack.


The similar system was described in article \cite{hua2018}, where blockchain was used to store information about the battery life cycle and solve the problem of its replacement for a transaction without trust between the parties. Thus, the calculation of the battery price and the exchange of digital currency between an EV owner and the station, as well as the key logic, were implemented using smart contracts to solve the problem of lack of trust. Similarly, the study \cite{strugar2019architecture} introduced an autonomous charging architecture and a billing framework for EV charging based on IOTA technology that provides resistance to hacker attacks and preserving confidential user information.


\subsection{Ensuring the Integrity and Security based on Blockchain}
The one of the key features of the multi-agent robotic systems is the increased requirements to the security in conditions of limited energy consumption of the devices, high requirements to productivity, physical characteristics and mobility of devices, their compatibility with each other. The presence of both software and hardware/software components interacting with each other in multi-agent robotic systems and their environment as well as possible variability of cyber-physical environment determine the susceptibility of such systems to specific sets of attacks.

To meet today's challenges, it is necessary to develop an integrated approach for security \cite{desnitsky2016} of robotic systems. The comprehensiveness of the approach here means not only the union of various security systems and it is also very important to take into account the protection of the security system in itself against the attacks. And here the blockchain can be very effective solution because of its reliability.

There are three basic security concepts - confidentiality, integrity, and availability. Cryptographic protocols, encrypted storage, etc. are usually used to ensure the confidentiality. For the availability support the intrusion detection and prevention systems, backup communication links, etc. are usually used. To ensure the integrity, checksum or digital signature are widely used. However, the use of blockchain technology can significantly improve the efficiency of ensuring the integrity of stored and transmitted data.

Let's consider the main areas of the distributed multi-agent robotic systems where the blockchain can be used \cite{ferrer2018blockchain}. 
\begin{itemize}
\item Robot's sensors \cite{moinet2017};
\item Robot's storage \cite{shafagh2017};
\item Robot's architecture \cite{desnitsky2012};
\item Interaction of agent with neighboring agents by IoT protocols \cite{reilly2019};
\item Interaction of agent with distant hosts or cloud through Internet \cite{fotiou2018};
\item Data aggregation, analysis and storage in the cloud \cite{xu2018}.
\end{itemize}

As an example, to enhance the security of the multi-agent robotic systems interaction, the new possible architecture of internet - Named Data Networking (NDN) can be used as a part of the system. The NDN is formed with two basic things, i.e. Sending Request and Receiving the data packet. Regarding the aspect of securities issues NDN can use approaches along with the blockchain to protect itself against the various threats.

There are many types of attacks to be noted \cite{abdallah2015}, e.g. Interest Flooding, Cache Misappropriation, Data Fishing, Selfish Attack etc. It is hard to avoid these security attacks by existing security solutions due to the decentralized and dynamic characteristics of NDN. But the decentralized blockchain \cite{sharma2017} approach can be applied to meet the security requirement of NDN. It can use the hash of Interest or Data by smart contract and it will decrease the chance of user privacy leaking because both data identifier and user identifier will be replaced with temporary names. In addition, nobody will be able to change or delete the fields due to the blockchain structure of hashed chain blockchain \cite{kshetri2017}.

It is already proven \cite{zhu2018security} that malicious actions can be detected by NDN blockchain model and blockchain with smart contract plays an efficient role in this scenario. The blockchain maintains the trajectory of Interest and Data in a hashed manner. The distributed blockchain for NDN contain series of blocks, and each block contains a hashed transition set. Each block has a head pointer (except the initiated block) that linked to a previous block. It consists of a timestamp record the time when the block is written, a bit that linked to a successor block. In each block, it keeps a hash value that composed by several hashed Interest or Data transition records.
\section{Realization of Multi-Agent Robotic System via Wireless Sensor Network}
\label{wsn}

Wireless sensor networks (WSN) are widely used for the practical implementation of multi-agent systems, and the addition of mobile robots to the WSN structure is a well-observed trend \cite{ahmed2017robotized}. Robots that are active doers in a multi-agent robotic system (MARS) can provide flexibility when installing network sensors and realizing active data acquisition, since they can perform various operations and interact with the environment \cite{ahmed2017robotized, afanasyev2019towards}. Although these interactions can be predetermined or based on real-time observations, however, the choice of a suitable communication protocol for robotized WSN can be a challenge, considering the complexity and multi-components of robots, as well as the type of communication implemented in MARS: one-to-many. One possible solution is to use the HTTP protocol over a WiFi connection, although it is not very suitable for bidirectional communication due to such difficulties as specifying ports and sometimes IP addresses for each network component, large packet size, high power consumption, and transmission problems for control commands via the Internet connection \cite{ahmed2017robotized}. The alternative solution is to use Cloud Computing and the Internet of Things (IoT) technologies for organizing communication between nodes and controlling the WSN robotized components, especially when using the Message Queue Telemetry Transport (MQTT) protocol. Due to small packets' size and "publish/subscribe" concept, managing the connection between network devices can be simpler and more feasible. 

Thus, at present 6LoWPAN networks with MQTT protocol are becoming a good solution for MARS applications due to low power consumption, IP-driven nodes and support for large mesh networks.
~
As an example, the studies \cite{ahmed2017robotized, ahmed2018environmental} proposed a robotized WSN suitable for solving various problems of environmental monitoring. The robotized WSN is an adaptive system in which intelligent agents are moving sensors for detecting and tracking areas, where the monitored environment parameters are different from certain threshold values. For the experiment, iRobot Create and KUKA youBot mobile platforms were used with additionally installed single-chip Gumstix Verdex pro TM XL6P computers and various expansion modules as robotic agents (mobile network nodes). The functionality of the proposed WSN robotic with data exchange using 6LoWPAN is verified using the MQTTBox platform. It enables building MQTT clients for publishing or subscribing topics, configuring MQTT virtual device networks, testing MQTT devices etc.

However, security becomes a critical issue when applying IoT concept to organize communication between the robotized WSN nodes. A successful solution of the problem of interaction between the nodes, together with recording the interaction history and performing the verification task can be provided by the blockchain technology (see, the Section \ref{classification} \textit{H}). This can increase the efficiency of the robotized WSN and expand the possibilities of their applications.

\begin{figure}[!htbp]
    \centering
    \includegraphics[width=0.4\textwidth]{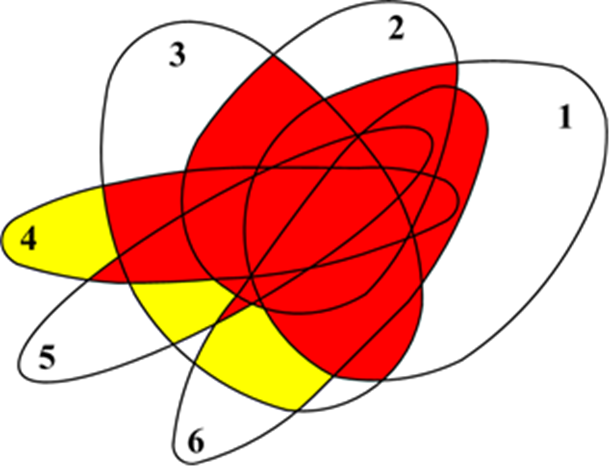}
    \caption{An ontology-related preprocessing scheme with six security-based ontologies and their combinations that identify dangerous processing nodes (red), warning zones (yellow), and safe (uncolored) zones}
    \label{fig02}
\end{figure}

The notion of a blockchain is associated with a publicly available chronological database of transactions recorded by a network of agents (e.g. swarm of agents). It is obvious that software robots (holons) are suitable for this purpose and such a research with robotic swarms had been executed \cite{ferrer2018blockchain}. Each agent possesses private and public keys used to prove the origin and/or encrypt messages. As stated in \cite{ferrer2018blockchain}, the proof may delay the task execution up to 10 minutes. In order to improve this situation, it is reasonable to use other cryptocurrencies instead of Bitcoin, for example, based on IOTA or Ethereum. On the other hand, not all transactions should be included in blocks. In many cases, different preprocessing schemes should be applied to deeply model the environment and classify the situation \cite{jotsov2016proposals}. Such an example is shown in Figure \ref{fig02}, where six security-based ontologies have been depicted and their combination defines dangerous processing nodes (in red), warning zones (in yellow), and safe (uncolored) zones. Blockchain-based solutions are required only in red and sometimes in yellow zones. The information preprocessing allows not only increasing the efficiency of the system, but also reducing its vulnerability. Other applications for modeling and data processing, also suitable for blockchain agents, are discussed in \cite{jotsov2016proposals}.

\section{Smart Buildings, Smart Cities and Industry 4.0}
\label{smart_cities}

The increasing integration of devices and robots with the Internet infrastructure and everyday activities is leading to a vision of the future where the Internet itself is progressively disappearing from view, i.e. the conscious actions necessary to connect to the network and transfer packets of data on it are becoming more and more transparent. The future citizen will not refer to the \textit{Internet} as often as we do in the same way we now simply connect devices in a stable manner very differently from what we used to do two decades ago, when more conscious effort was necessary and the connection could be lost several times over a work session. The more the Internet will enter our life, the less we will notice, the more pervasive it will become in every aspect of personal and professional life. For example, the use of a multi-agent system with auditable blockchain voting helps to make the voting records transparent and unchangeable, allowing to overcome the main e-voting problem how to increase the level of respondents' trust in the electronic voting system \cite{Pawlak2018}.

This progressive integration is what today we call the \textit{Internet of Things} (IoT) \cite{alam2017internet}, where every object is transparently connected to the network and can communicate with other objects, systems or individuals. In this developing scenario robots also play a role, and robotics as a discipline cannot be considered as a completely separate domain and independently developing. The \textit{Internet of Robotics Things} (IoRT) \cite{ray2016internet, simoens2018internet, batth2018internet, mahieu2019semantics, afanasyev2019towards} is a recently defined concept, which aims at describing the integration of robotics technologies in IoT scenarios. Multi-Agent Robotic Systems, as described in this paper, is a notable application scenario for which IoRT can constitute the working infrastructure \cite{gautam2012review}, both for domestic (domotics/Smart Buildings) and industrial (Industry 4.0) use.

The papers \cite{casino2018, Wang2019} examine the integration of the blockchain technology and the Internet of Things, which is expected to transform human life and provide great economic benefits. However, the main restrictions for such integration are insufficient data security and a level of trust.


Regarding Smart Buildings and Software-defined Buildings (SDB), blockchain is destined to represent a persistence infrastructure of pervasive application in everything concerning home resource, environment and processes \cite{mazzara2019reference}, from energy management to billing, from environmental comfort to safety, surveillance and further. Developing the idea of Smart Building system, researchers integrate subsystems, such as intelligent networks, services, buildings and household appliances, into models of Smart Spaces and even Smart Cities, using blockchains to effectively exchange data when interacting subsystems, connecting and remote control for reaching a better life quality, sustainability, energy conservation and the development of socio-economic systems \cite{lazaroiu2017, mazzara2019reference}. The IoRT in the domotics context will be the enabling infrastructure for blockchain-based Multi-Agent Robotic Systems. In this context, and it also applies to Smart Cities as aggregation of Smart Buildings, a large number of sensors collect data with high variability of accuracy, reliability and frequency. Therefore, the Blockchain technology could permit the management of public immutable ledgers tracking all the activities and determining those that are more trustful and those that are less, and act accordingly. This hardware and software infrastructure will simplify the interaction of the different agents in the different buildings allowing traceability of data collection and enhancing trust, security and accuracy of the cooperation between multiple agents.

Ecological and environmental monitoring is a sphere where all advantages of blockchain are needed, and usage of IoT and IoRT is a good way to create a big independent sensor network. Transparency, immutability and security are highly demanded for environmental monitoring. Peer-to-peer approach gives a way of a cheap connection of the new sensor to the global network and start provide a data about environment publicly. Right now the concept of citizens' observatories one of the most suitable for blockchain technology. Such projects as WeObserve \cite{berti2019not} and WeSenseIt \cite{lanfranchi2014citizens} demonstrate how citizens sensor networks helping to improve fullness of the ecological information, and with peering technologies it can be a really scalable solution.

In the context of Industry 4.0 (Smart Factories) \cite{Forbes16}, and with the increasing trend of automation, similar considerations on the efficacy of IoRT and Blockchain hold. Blockchain technology may eventually represent the pillar of a business or organization thanks to better contract management, effective quality control, better accountability, recognition and authentication of IoRT devices. In general, blockchain technology in Industry 4.0 gives us the chance to innovate and refresh the concept of cybersecurity, offering a mechanism by which activities can be immutably tracked and pseudonymized \cite{fernandez2019review}.

The progress in various technologies and their cooperation for robotics, automation, IoT, big data processing, cloud computing and blockchain lead to the fourth industrial revolution, when the interaction of the Smart Factory components within the company and external industrial IoT systems provide trust and reliable control over the resource distribution and products \cite{teslya2017, kapitonov2018}.
For example, in logistics, where the supply chain is a multi-agent system in which each supplier has own behavior model and purposes, the blockchain can bring the necessary transparency and trust, speeding up the supply processes and eliminating many shortcomings of current supply chains \cite{casino2018, casado2018blockchain}.
\section{Conclusions and Discussion}
\label{discussion}

Currently, the approach consisting in the organization of an immutable distributed database storing all relevant information and providing access to agents of a multi-agent robotic system (therefore expanding the capabilities of the system as a whole) is of great relevance in the context of the Fourth Industrial Revolution. The key component of the approach is a distributed ledger technology (blockchain), which allows agents to interact or allocate the tasks through responsible “smart contracts”. On the one hand, the reliability of the agent is mainly determined by the reputational model, allowing to determine the trust level to the agent only after the fulfillment of the agreed obligations. On the other hand, the automation of obligation fulfillment by an agent can provide a verification procedure that will allow to verify the liability execution even among initially unknown agents and reach a consensus in conditions without trust (\cite{ferrer2018blockchain, danilov2018, Skowronski2019}). In addition, the goals for the blockchain implementation in a multi-agent robotic system may be the increase of the interaction efficiency between agents by organizing more trusted information support, assessing the robot productivity or detecting performance problems, voting consensus for swarm robotics, plan scheduling and task allocation, deploying distributed decision making and collaborative missions.

The basic message of this paper is to bring together and provide, as a guide, ideas on how frameworks, architectures and structures supported by blockchain solutions can be used to solve practical problems that face by multi-agent robotic systems and cyber-physical systems. Relevant studies show that the blockchain begins to play a large role in the development of systems and applications with many agents (robots), in which the development of strategies for their coordination is conducted in such a way that agents can effectively perform their operations and intelligently coordinate task allocation \cite{basegio2017decentralised}. The analysis of the recent publications allowed to identify and classify groups of tasks for multi-agent robotic systems based on blockchain technology. This classification is our main contribution by this paper. Blockchain technologies can be used to expand the existing number of platforms and libraries used by researchers, or to motivate them to use a common solution that is widely distributed and tested, rather than trying to develop their own software solutions to cover similar scenarios.
Real-world scenarios may require the use of disparate agents and the performance of tasks with different structures, constraints and complexity. Therefore, the requirements for the quality of communication in decentralized systems are increasing, including such important functions as maintaining resiliency, data integrity and security in accessing data. Therefore, the introduction of blockchain technology for the interaction and coordination of multi-agent robotic systems becomes a reasonable solution for many modern research and industrial tasks.

 
Based on modern investigations, the authors conclude that at present one of the most promising tasks in the field of developing multi-agent systems is the development of methodologies, models, structures, architectures and methods, aiming to integrate blockchain technology with high complexity systems, such as cyber-physical systems, robot swarm, the Internet of things, the Internet of Robotic Things, Smart Buildings, Smart Factories and Smart Cities. To fully implement such integration, it is necessary to automate and solve subtasks of different difficulty levels, such as intellectual support for agent interaction, task and plan allocation, analysis of task performance and liability execution, evaluation of agent performance,  identification of  improperly functioning agents and intruders, security-based issues and many others.

Many tasks remain to be done at the moment and require solutions, but some of them, according to the authors, are the most important and urgent \cite{afanasyev2019blockchain}:
\begin{itemize}
\item Development of a conceptual model of information support for robot network during the task performance; 
\item Design of a typical ontological model of a multi-robot network; 
\item Development of requirements and formal process modeling for liability execution \cite{Mazzara2010, danilov2018};
\item Design of a consensus protocol for a group interaction verification before launching a task based on the information from a distributed ledger;
\item Development of a validation methodology for task performance by the robotic system; 
\item Design of multi-agent robotic system architecture; 
\item Automatic process reconfiguration for multi-robot systems based on real scenarios;
\item Analysis of cybersecurity \cite{dragoni2016internet};
\item Improvement of existing frameworks allowing for multi-agent robotic networks to perform collaborative tasks with regard to scalability, decentralization and security requirements.
\end{itemize}

\bibliographystyle{ieeetr}
\bibliography{bibliography}

\begin{thebibliography}{10}

\bibitem{ray2016internet}
P.~P. Ray, ``Internet of robotic things: concept, technologies, and
  challenges,'' {\em IEEE Access}, vol.~4, pp.~9489--9500, 2016.

\bibitem{simoens2018internet}
P.~Simoens, M.~Dragone, and A.~Saffiotti, ``The internet of robotic things: A
  review of concept, added value and applications,'' {\em International Journal
  of Advanced Robotic Systems}, vol.~15, no.~1, 2018.

\bibitem{batth2018internet}
R.~S. Batth, A.~Nayyar, and A.~Nagpal, ``Internet of robotic things: Driving
  intelligent robotics of future-concept, architecture, applications and
  technologies,'' in {\em 2018 4th International Conference on Computing
  Sciences (ICCS)}, pp.~151--160, IEEE, 2018.

\bibitem{mahieu2019semantics}
C.~Mahieu, F.~Ongenae, F.~De~Backere, P.~Bonte, F.~De~Turck, and P.~Simoens,
  ``Semantics-based platform for context-aware and personalized robot
  interaction in the internet of robotic things,'' {\em Journal of Systems and
  Software}, vol.~149, pp.~138--157, 2019.

\bibitem{afanasyev2019towards}
I.~Afanasyev, M.~Mazzara, S.~Chakraborty, N.~Zhuchkov, A.~Maksatbek, M.~Kassab,
  and S.~Distefano, ``Towards the internet of robotic things: Analysis,
  architecture, components and challenges,'' {\em arXiv preprint
  arXiv:1907.03817}, 2019.

\bibitem{kapitonov2017}
A.~Kapitonov, S.~Lonshakov, A.~Krupenkin, and I.~Berman, ``Blockchain-based
  protocol of autonomous business activity for multi-agent systems consisting
  of uavs,'' in {\em Workshop on Research, Education and Development of
  Unmanned Aerial Systems (RED-UAS)}, pp.~84--89, IEEE, 2017.

\bibitem{ferrer2018blockchain}
E.~C. Ferrer, ``The blockchain: a new framework for robotic swarm systems,'' in
  {\em Proceedings of the Future Technologies Conference}, pp.~1037--1058,
  Springer, 2018.

\bibitem{strobel2018managing}
V.~Strobel, E.~Castell{\'o}~Ferrer, and M.~Dorigo, ``Managing byzantine robots
  via blockchain technology in a swarm robotics collective decision making
  scenario,'' in {\em Proceedings of the 17th International Conference on
  Autonomous Agents and MultiAgent Systems}, pp.~541--549, International
  Foundation for Autonomous Agents and Multiagent Systems, 2018.

\bibitem{afanasyev2019blockchain}
I.~Afanasyev, A.~Kolotov, R.~Rezin, K.~Danilov, A.~Kashevnik, and V.~Jotsov,
  ``Blockchain solutions for multi-agent robotic systems: Related work and open
  questions,'' in {\em Proceedings of the 24th Conference of Open Innovations
  Association FRUCT}, p.~76, FRUCT Oy, 2019.

\bibitem{basegio2017decentralised}
T.~L. Basegio, R.~A. Michelin, A.~F. Zorzo, and R.~H. Bordini, ``A
  decentralised approach to task allocation using blockchain,'' in {\em
  International Workshop on Engineering Multi-Agent Systems}, pp.~75--91,
  Springer, 2017.

\bibitem{edChain}
``Blockchain faq.'' \url{https://medium.com/edchain}.
\newblock Accessed: 2019-07-07.

\bibitem{coinbureau2019}
``Solving the blockchain trilemma: Decentralization, security \& scalability.''
  \url{https://www.coinbureau.com/analysis/solving-blockchain-trilemma/}.
\newblock Accessed: 2019-07-07.

\bibitem{andoni2019blockchain}
M.~Andoni, V.~Robu, D.~Flynn, S.~Abram, D.~Geach, D.~Jenkins, P.~McCallum, and
  A.~Peacock, ``Blockchain technology in the energy sector: A systematic review
  of challenges and opportunities,'' {\em Renewable and Sustainable Energy
  Reviews}, vol.~100, pp.~143--174, 2019.

\bibitem{zikratov2016dynamic}
I.~Zikratov, O.~Maslennikov, I.~Lebedev, A.~Ometov, and S.~Andreev, ``Dynamic
  trust management framework for robotic multi-agent systems,'' in {\em
  Internet of Things, Smart Spaces, and Next Generation Networks and Systems},
  pp.~339--348, Springer, 2016.

\bibitem{danilov2018}
K.~Danilov, R.~Rezin, A.~Kolotov, and I.~Afanasyev, ``Towards blockchain-based
  robonomics: autonomous agents behavior validation,'' in {\em International
  Conference on Intelligent Systems}, IEEE, 2018.

\bibitem{teslya2018}
N.~Teslya and A.~Smirnov, ``Blockchain-based framework for ontology-oriented
  robots’ coalition formation in cyberphysical systems,'' in {\em MATEC Web
  of Conferences}, vol.~161, p.~03018, EDP Sciences, 2018.

\bibitem{Kashevnik2018}
A.~Kashevnik and N.~Teslya, ``{Blockchain-Oriented Coalition Formation by CPS
  Resources: Ontological Approach and Case Study},'' {\em Electronics}, vol.~7,
  p.~66, may 2018.

\bibitem{lopes2019}
V.~Lopes, L.~A. Alexandre, and N.~Pereira, ``Controlling robots using
  artificial intelligence and a consortium blockchain,'' {\em arXiv preprint
  arXiv:1903.00660}, 2019.

\bibitem{ferrer2018robochain}
E.~C. Ferrer, O.~Rudovic, T.~Hardjono, and A.~Pentland, ``Robochain: A secure
  data-sharing framework for human-robot interaction,'' {\em arXiv preprint
  arXiv:1802.04480}, 2018.

\bibitem{shukla2018}
A.~Shukla, S.~K. Mohalik, and R.~Badrinath, ``Smart contracts for multiagent
  plan execution in untrusted cyber-physical systems,'' in {\em 2018 IEEE 25th
  International Conference on High Performance Computing Workshops (HiPCW)},
  pp.~86--94, IEEE, 2018.

\bibitem{Skowronski2019}
R.~Skowro{\'{n}}ski, ``{The open blockchain-aided multi-agent symbiotic
  cyber–physical systems},'' {\em Future Generation Computer Systems},
  vol.~94, pp.~430--443, may 2019.

\bibitem{zlot2002}
R.~Zlot, A.~Stentz, M.~B. Dias, and S.~Thayer, ``Multi-robot exploration
  controlled by a market economy,'' in {\em Robotics and Automation, 2002.
  Proceedings. ICRA'02. IEEE International Conference on}, vol.~3,
  pp.~3016--3023, IEEE, 2002.

\bibitem{hua2018}
S.~Hua, E.~Zhou, B.~Pi, J.~Sun, Y.~Nomura, and H.~Kurihara, ``Apply blockchain
  technology to electric vehicle battery refueling,'' in {\em Proceedings of
  the 51st International Conference on System Sciences}, 2018.

\bibitem{strugar2019architecture}
D.~Strugar, R.~Hussain, M.~Mazzara, V.~Rivera, I.~Afanasyev, and J.~Lee, ``An
  architecture for distributed ledger-based m2m auditing for electric
  autonomous vehicles,'' in {\em Workshops of the International Conference on
  Advanced Information Networking and Applications}, pp.~116--128, Springer,
  2019.

\bibitem{desnitsky2016}
V.~Desnitsky, A.~Chechulin, I.~Kotenko, D.~Levshun, and M.~Kolomeec,
  ``Application of a technique for secure embedded device design based on
  combining security components for creation of a perimeter protection
  system,'' pp.~609--616, 2016.

\bibitem{moinet2017}
A.~Moinet, B.~Darties, and J.-L. Baril, ``Blockchain based trust \&
  authentication for decentralized sensor networks,'' 06 2017.

\bibitem{shafagh2017}
H.~Shafagh, L.~Burkhalter, A.~Hithnawi, and S.~Duquennoy, ``Towards
  blockchain-based auditable storage and sharing of iot data,'' in {\em
  Proceedings of the 2017 on Cloud Computing Security Workshop}, pp.~45--50,
  ACM, 2017.

\bibitem{desnitsky2012}
V.~Desnitsky, I.~Kotenko, and A.~Chechulin, ``Configuration-based approach to
  embedded device security,'' {\em Lecture Notes in Computer Science (including
  subseries Lecture Notes in Artificial Intelligence and Lecture Notes in
  Bioinformatics)}, vol.~7531 LNCS, pp.~270--285, 2012.

\bibitem{reilly2019}
E.~E. Reilly, M.~M. Maloney, M.~Siegel, and G.~Falco, ``A smart city iot
  integrity-first communication protocol via an ethereum blockchain light
  client,'' in {\em Proceedings of the International Workshop on Software
  Engineering Research and Practices for the Internet of Things (SERP4IoT
  2019), Marrakech, Morocco}, 2019.

\bibitem{fotiou2018}
N.~Fotiou, V.~Siris, and G.~Polyzos, ``Interacting with the internet of things
  using smart contracts and blockchain technologies,'' in {\em Security,
  Privacy, and Anonymity in Computation, Communication, and Storage},
  pp.~443--452, Springer International Publishin, 2018.

\bibitem{xu2018}
Q.~Xu, K.~Aung, Y.~Zhu, and K.~Leong~Yong, {\em A Blockchain-Based Storage
  System for Data Analytics in the Internet of Things}, pp.~119--138.
\newblock 06 2018.

\bibitem{abdallah2015}
E.~AbdAllah, H.~Hassanein, and M.~Zulkernine, ``A survey of security attacks in
  information-centric networking,'' {\em IEEE Communications Surveys
  Tutorials}, vol.~17, no.~3, pp.~1441--1454, 2015.

\bibitem{sharma2017}
P.~K. Sharma, S.~Singh, Y.-S. Jeong, and J.~H. Park, ``Distblocknet: A
  distributed blockchains-based secure sdn architecture for iot networks,''
  {\em IEEE Communications Magazine}, vol.~55, no.~9, pp.~78--85, 2017.

\bibitem{kshetri2017}
N.~Kshetri, ``Can blockchain strengthen the internet of things?,'' {\em IT
  Professional}, vol.~19, no.~4, pp.~68--72, 2017.

\bibitem{zhu2018security}
K.~Zhu, Z.~Chen, W.~Yan, and L.~Zhang, ``Security attacks in named data
  networking of things and a blockchain solution,'' {\em IEEE Internet of
  Things Journal}, 2018.

\bibitem{ahmed2017robotized}
S.~Ahmed, A.~Topalov, and N.~Shakev, ``A robotized wireless sensor network
  based on mqtt cloud computing,'' in {\em 2017 IEEE International Workshop of
  Electronics, Control, Measurement, Signals and their Application to
  Mechatronics (ECMSM)}, pp.~1--6, IEEE, 2017.

\bibitem{ahmed2018environmental}
S.~A. Ahmed, V.~L. Popov, A.~V. Topalov, and N.~G. Shakev, ``Environmental
  monitoring using a robotized wireless sensor network,'' {\em AI \& SOCIETY},
  vol.~33, no.~2, pp.~207--214, 2018.

\bibitem{jotsov2016proposals}
V.~S. Jotsov, ``Proposals for knowledge driven and data driven applications in
  security systems,'' in {\em Innovative Issues in Intelligent Systems},
  pp.~231--293, Springer, 2016.

\bibitem{Pawlak2018}
M.~Pawlak, A.~Poniszewska-Mara{\'{n}}da, and N.~Kryvinska, ``{Towards the
  intelligent agents for blockchain e-voting system},'' {\em Procedia Computer
  Science}, vol.~141, pp.~239--246, jan 2018.

\bibitem{alam2017internet}
N.~Alam, P.~Vats, and N.~Kashyap, ``Internet of things: A literature review,''
  in {\em 2017 Recent Developments in Control, Automation \& Power Engineering
  (RDCAPE)}, pp.~192--197, IEEE, 2017.

\bibitem{gautam2012review}
A.~Gautam and S.~Mohan, ``A review of research in multi-robot systems,'' in
  {\em 2012 IEEE 7th International Conference on Industrial and Information
  Systems (ICIIS)}, pp.~1--5, IEEE, 2012.

\bibitem{casino2018}
F.~Casino, T.~K. Dasaklis, and C.~Patsakis, ``A systematic literature review of
  blockchain-based applications: current status, classification and open
  issues,'' {\em Telematics and Informatics}, 2018.

\bibitem{Wang2019}
X.~Wang, X.~Zha, W.~Ni, R.~P. Liu, Y.~J. Guo, X.~Niu, and K.~Zheng, ``{Survey
  on blockchain for Internet of Things},'' {\em Computer Communications},
  vol.~136, pp.~10--29, feb 2019.

\bibitem{mazzara2019reference}
M.~Mazzara, I.~Afanasyev, S.~R. Sarangi, S.~Distefano, and V.~Kumar, ``A
  reference architecture for smart and software-defined buildings,'' {\em arXiv
  preprint arXiv:1902.09464}, 2019.

\bibitem{lazaroiu2017}
C.~Lazaroiu and M.~Roscia, ``Smart district through iot and blockchain,'' in
  {\em 2017 IEEE 6th International Conference on Renewable Energy Research and
  Applications (ICRERA)}, pp.~454--461, IEEE, 2017.

\bibitem{berti2019not}
A.~Berti~Suman and M.~Van~Geenhuizen, ``Not just noise monitoring: rethinking
  citizen sensing for risk-related problem-solving,'' {\em Journal of
  Environmental Planning and Management}, pp.~1--22, 2019.

\bibitem{lanfranchi2014citizens}
V.~Lanfranchi, S.~N. Wrigley, N.~Ireson, U.~Wehn, and F.~Ciravegna, ``Citizens'
  observatories for situation awareness in flooding,'' in {\em ISCRAM 2014
  Conference Proceedings-11th International Conference on Information Systems
  for Crisis Response and Management}, pp.~145--154, Sheffield, 2014.

\bibitem{Forbes16}
Forbes, ``Why everyone must get ready for the 4th industrial revolution.''
  \url{https://www.forbes.com/sites/bernardmarr/2016/04/05/why-everyone-must-get-ready-for-4th-industrial-revolution/#69263d403f90},
  2016.
\newblock Accessed: 2019-07-07.

\bibitem{fernandez2019review}
T.~M. Fern{\'a}ndez-Caram{\'e}s and P.~Fraga-Lamas, ``A review on the
  application of blockchain to the next generation of cybersecure industry 4.0
  smart factories,'' {\em IEEE Access}, vol.~7, pp.~45201--45218, 2019.

\bibitem{teslya2017}
N.~Teslya and I.~Ryabchikov, ``Blockchain-based platform architecture for
  industrial iot,'' in {\em 2017 21st Conference of Open Innovations
  Association (FRUCT)}, pp.~321--329, IEEE, 2017.

\bibitem{kapitonov2018}
A.~Kapitonov, I.~Berman, S.~Lonshakov, and A.~Krupenkin, ``Blockchain based
  protocol for economical communication in industry 4.0,'' in {\em 2018 Crypto
  Valley Conference on Blockchain Technology (CVCBT)}, pp.~41--44, IEEE, 2018.

\bibitem{casado2018blockchain}
R.~Casado-Vara, J.~Prieto, F.~De~la Prieta, and J.~M. Corchado, ``How
  blockchain improves the supply chain: Case study alimentary supply chain,''
  {\em Procedia computer science}, vol.~134, pp.~393--398, 2018.

\bibitem{Mazzara2010}
M.~Mazzara, ``Deriving specifications of dependable systems: toward a method,''
  {\em CoRR}, vol.~abs/1009.3911, 2010.

\bibitem{dragoni2016internet}
N.~Dragoni, A.~Giaretta, and M.~Mazzara, ``The internet of hackable things,''
  in {\em International Conference in Software Engineering for Defence
  Applications}, pp.~129--140, Springer, 2016.

\end{thebibliography}

\end{document}